# Conceptual Content in Deep Convolutional Neural Networks: An Analysis into Multi-faceted Properties of Neurons

Zahra Sadeghi

Computer Vision Center, Universitat Autònoma de Barcelona
zahsade@gmail.com

**Abstract.** In this paper, convolutional layers of pre-trained VGG16 model are analyzed. The analysis is based on the responses of neurons to the images of classes in ImageNet database. First, a visualization method is proposed in order to illustrate the learned content of each neuron. Next, single and multi-faceted neurons are investigated based on the diversity of neurons responses to different category of objects. Finally, neuronal similarities at each layer are computed and compared. The results demonstrate that the neurons in lower layers exhibit a multi-faceted behavior, whereas the majority of neurons in higher layers comprise single-faceted property and tend to respond to a smaller number of concepts.

**Keywords:** Convolutional Deep Neural Networks, Deep Visualization, Multi-faceted and single-faceted Neurons

## 1    Introduction

The non-linear deep learning approach has achieved a tremendous success in various applications and has revolutionized the area of machine learning and computer vision in recent years. In particular, convolutional neural networks have demonstrated a high flexibility and efficiency that can be utilized for a wide range of problems in different domains. In spite of the increasing efforts in designing new deep network architectures, still not much is known about the internal representation of these networks and still the trained units of deep neural networks are not very interpretable. To date, two principle directions that has been developed for understanding deep neural networks are network analysis and network visualization. Several authors have attempted to analyze DNNs using information theory principles in order to study deep representation. In this regard, (Shwartz-Ziv & Tishby, 2017) (Tishby & Zaslavsky, 2015) viewed DNN as a Markovian compression procedure between input and output of the network and calculated the information bottleneck bound on each layer. In a follow-up research, (Achille & Soatto, 2017) worked on the minimal deep representation. Other researchers have focused on the problem of deep network visualization in order to gain an understanding about the underlying learning structure of these networks. Deep network understanding approaches can be divided into two broad categories which follow macroscopic or micro-scopic viewpoints. In the first category, the primary interest is centered around understanding the functionality of deep neural networks as a classification tool and the related questions are about the effect of each pixel or group of pixels in a patch of images on the performance of the fully connected layers of network in making discrimination between classes. In contrast, the second category seeks to inspect the building blocks of pre-trained deep networks such as neurons and weights. Regarding the former view, Simonyan et al., conducted a sensitivity analysis and created a saliency map based on the influence of input image pixels on the classification decision (Simonyan & Zisserman, 2014). In a similar vein, Zintgraf et al., studied the effect of each patch that contains a specific pixel (Zintgraf, Cohen, & Welling, 2016). Zhou et al., derived a minimal representation of input images by removing the regions that caused the least decrement in correct classification rate

 (Zhou, Lapedriza, Xiao, Torralba, & Oliva, 2014)(Zhou, Khosla, Lapedriza, Oliva, & Torralba, 2014). On the latter view, the idea of activation maximization has been followed by several researchers. In a number of visualization studies, authors paid attention to the maximum activation values in convolutional and pooling layers and retrieved the corresponding input patches of images(Girshick, Donahue, Darrell, & Malik, 2014)(Erhan, Bengio, Courville, & Vincent, 2009). Yosinsky et al., applied a regularization technique in order to find images that cause highest activation values (Yosinski, Clune, Nguyen, Fuchs, & Lipson, 2015). Zeiler and Fergus introduced the idea of deconvolution in which they employed reverse computation from activation maps to the input patterns to obtain a heat map that visualizes the most effective area of image that is remarkable in provoking a neuron to fire (Zeiler & Fergus, 2014).

One intriguing property that has been discovered about neurons is that they can respond to various concepts. These types of neurons which are called multi-faceted neurons have been found in mediotemporal lobe of the brain and have reported to be activated by pictures of multiple objects (Quiroga, Reddy, Kreiman, Koch, & Fried, 2005). In the realm of deep visualization, a few efforts have been made to identify these neurons in deep neural networks. Nguyen et al., focused on the problem of multi-faceted visualization (Nguyen, Yosinski, & Clune, 2016).  They used an optimization algorithm to produce images that maximally activate each neuron. In order to create multi-faceted visualization, they used different initializations using images that were obtained from cluster of images from each class. Along the same lines with the recent findings in deep network interpretation, in this paper, a number of statistical analytical tools are conducted with the purpose of scrutinizing the trained neurons and uncovering the progressive patterns of learning in the deep learning pipeline. In this work, it is hypothesized that the association between neurons can be derived based on the category of object that can activate them and it encompasses conceptual information about the deep layers' representation. In particular, the line of this work is concerned with multi-faceted properties of neurons in terms of quantification of the degree of neurons' responses to different classes as well as a neuron visualization method based on independent components of the top images that activate each neuron. Moreover, conceptual similarity between neurons across deep convolutional layers is investigated. In the following sec-tions, first the conceptual encoding of neurons is described. Next, a method for visualization of content of neurons is proposed. Then multi-faceted and single-faceted neurons are defined. Finally, the neuronal similarity matrices are presented and discussed.

## 2 Method

### 2.1 Conceptual Encoding

The goal of this paper is to probe the learned concept which is developed across deep convolutional layers. To this end, VGG16 convolutional neural network, which is pre-trained on ImageNet2012 dataset is studied. This dataset contains 1.3 million images from 1000 category of objects. VGG16 convolutional neural network includes five blocks of convolutional layers. Table 1 lists the number of convolutional layers and neurons in each block. Throughout the paper, deep layer numbers are also considered regardless of the block number. For example, layer 3 refers to the second convolutional layer in block 2.

**Table 1.** VGG16 net architecture

| Block #no.    | 1  | 2   | 3   | 4   | 5   |
|---------------|----|-----|-----|-----|-----|
| #Conv. Layers | 2  | 2   | 3   | 3   | 3   |
| #Neurons      | 64 | 128 | 256 | 512 | 512 |

In order to capture the association between neurons activations and the corresponding triggered classes, the following approach is taken. The conceptual content of each layer is encoded based on the appeared object classes in the top 100 list of images that achieved the highest amount of activation values (Girshick et al., 2014)(Erhan et al., 2009). For this purpose, a neuron-by-class co-occurrence matrix (or *CoF* matrix) is defined which counts the frequency of highly activated images. Hence, *CoF(n,c)* denotes the number of images that belongs to object class *c* which has highly provoked neuron *n*.

### 2.2 Neuron Visualization

In this section, a method for visualization of the information content of neurons is introduced. This method is based on extraction of basis patterns in 100-top patches of input images that highly activate neurons at each layer. In addition, the underlying basis of top images yields an insight about the multi-faceted effect of neurons. Past research on neuron visualization has been mainly focused on providing one visualization for each neuron. In the current paper, it is hypothesized that more than one visualization can provide a more detailed explanation of neurons contents. Following this idea, it is assumed that there are multiple sources of information that stimulates each neuron. For this, Independent Component Analysis (ICA) is leveraged. ICA is a powerful tool that can provide high order representation of images and has been shown to be suitable for dealing with the problem of blind source separation (Hyvärinen, Karhunen, & Oja, 2001). This method can also be regarded as a dimension reduction technique which can find the original sources of signals. It has been shown that ICA behavior is similar to simple cells in the brain. In essence, the structure of ICA bases has shown to resemble the receptive fields of simple cells in primary cortex (Caywood, Willmore, & Tolhurst, 2004). I applied ICA to the list of patches pertaining to each neuron and then visualized the ICA bases or Independent Components (ICs). ICA is an unsupervised algorithm but the number of components needs to be determined in advance. Different numbers are tested and it is considered that eight components include representative information. The basic components provide visual information about the nature of the images preferred by each neuron. IC visualization can reveal the multi-responsive feature of neurons based on the similarity of the patterns captured by each component. If all the components look alike, it signifies that a neuron is a single-faceted neuron, otherwise, it will be considered as a multi-faceted neuron. This notion is discussed in the next section.

## 3     Results

### 3.1    Multi-faceted vs. Single-Faceted Neurons

Single-faceted (SF) neurons are referred to those neurons in human MTL with selective responses to a single concept (Quiroga et al., 2005). In this paper, SF neurons are defined as those neurons which highly respond to a particular class of images. These types of neurons are also known as concept neurons or grandmother cells. On the other hand, multi-faceted (MF) neurons are considered as the neurons which are activated by images from a range of different classes. In order to assess neurons' conceptual properties, neurons' responses to classes are compared based on their representations through the rows of *CoF* matrix. For this purpose, signal *flatness* and *sparsity* of each row of *CoF* matrix is computed. The sparseness of each neuron in one layer is calculated using Gini index (Zonoobi, Kassim, & Venkatesh, 2011). I chose Gini index because of a number of advantages that it has over other indexes. It produces normalized values in the range of [0,1] and the values are not dependent on the size of the input vector. The sparsity index captures the frequency of responses of a neuron to multiple classes. Low values of this index is an indication of multi-faceted neurons. In order to evaluate the flatness of each row, the spectral flatness is measured using equations (1) (Madhu, 2009), Where $n_r^l$ is the row *r* of *CoF* matrix at layer *l* and $N^l$ is the number of neurons in layer *l*.

$$\dot{n}_r^l = \frac{n_r^l}{\sum_{i=0}^{N^l-1} n_r^l(i)}$$
$$\log_2(\text{Flatness}(n_r^l) + 1) = -\frac{1}{\log_2(N^l)} \sum_i \dot{n}_r^l(i) \log_2(\dot{n}_r^l(i)) \quad (1)$$

In order to prevent the undesired effect of zeros in this measurement, first add a small constant of 0.0000001 is added to all the elements of $n_r^l$. Therefore, the flatness values ranges between 0 and 1. Lower values of sparsity and higher values of flatness is an indication of multi-faceted neurons. The multi-faceted capacity of a neuron $n_i$ is called MF degree and is defined by using equation (2).

$$MF(n_i) = \frac{flatness(n_i)}{sparsity(n_i)} \quad (2)$$

Fig. 1 illustrates the normalized MF values for all neurons across all layers. It can be observed that the MF degree of neurons in shallow layers is higher than neurons in deeper layers. It is worthwhile to note that while there are too many neurons with high MF degrees, the values are not significantly high. More importantly, the max value of un-normalized MF degree is 0.63, and the mean value is 0.53.

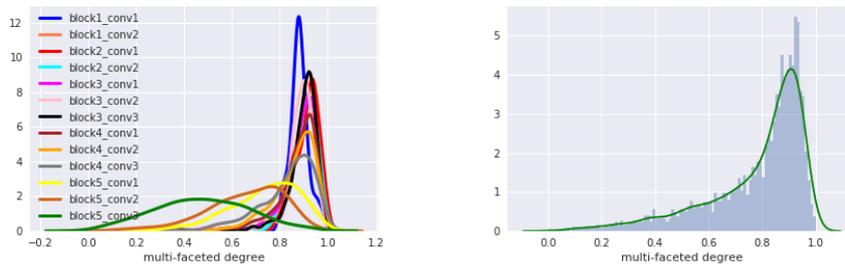

**Fig. 1.** Normal distribution of sparsity and flatness measurements corresponding to all of the neurons at each layer of vgg16 net

Furthermore, examples of visualization of one of the four-top neurons based on MF degrees are shown in Fig. 2 and Fig. 3. For each value, the corresponding p-value based on the distribution of all other MF values within the same layer is specified. Value of 0.05 is considered as the cut-off threshold for significance. The values above mean are assumed to indicate multi-faceted property. For each neuron, ICs are visualized in gray-scale as well as three different color configurations. The first column from left shows the ICs in gray scale. The second and third columns present two different methods (through using numpy.dstack and make_lupton_rgb) in which the ICs are computed individually for each R, G, and B channels and then the channels are combined into a three-dimensional array. The last column exhibits the converted results in 8-bit un-signed integer. In another attempt, first, all the channels are concatenated and then ICs are computed. However, since the visualizations were not semantically comprehensible they are not reported here. From the results, it can be understood that neurons with high MF degrees tend to have a blurred visualization and their corresponding ICs are dissimilar. In contrast, the visualization of neurons with low MF degrees contains clearer shapes. Fig. 4 shows examples of gray-scale visualization of MF and SF neurons that are selected in a qualitative manner along with their corresponding patches. It can be observed that ICs can visualize neurons that respond either to a specific concept or various concepts. High MF degree indicates that a neuron is activated by patches of images from different classes and it generally implies that the pattern of patches should be notably different. Furthermore, the t-test results suggests that unlike MF neurons, SF neurons cannot be found too often in shallower convolutional layers. Furthermore, the MF degree corresponding to top MF and SF neurons are depicted at each layer in Fig. 5. In addition, the number of MF and SF neurons scaled by the total number of neurons at each layer are shown in the same figure. As we can observe, while the MF degree of concept neurons keeps decreasing through the deep layers, it is quite constant for MF neurons. The results also provides evidence that the number of SF neurons grows across the deep layers, whereas the number of MF neurons presents a declining pattern.

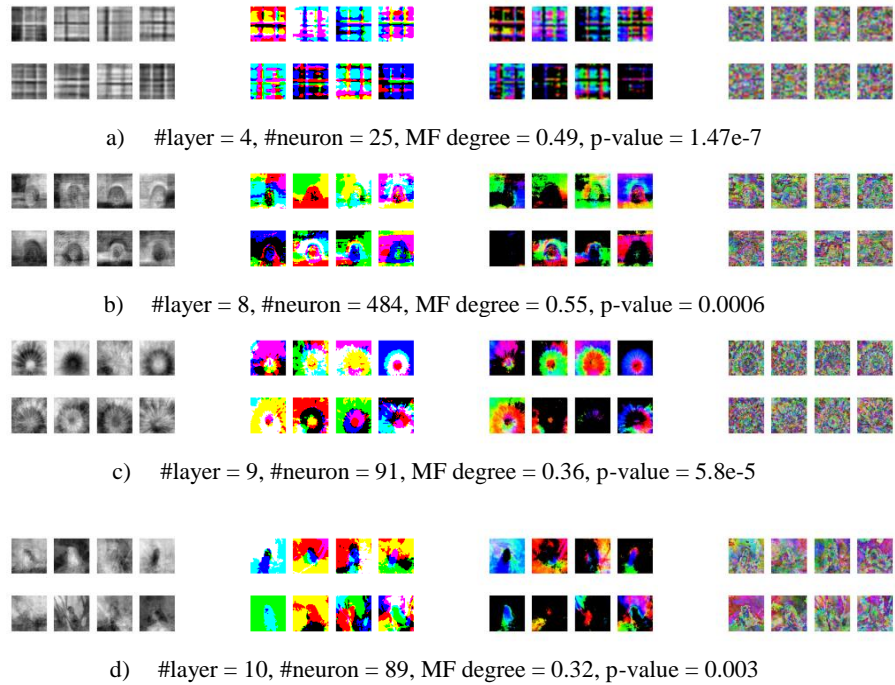

a)  #layer = 4, #neuron = 25, MF degree = 0.49, p-value = 1.47e-7

b)  #layer = 8, #neuron = 484, MF degree = 0.55, p-value = 0.0006

c)  #layer = 9, #neuron = 91, MF degree = 0.36, p-value = 5.8e-5

d)  #layer = 10, #neuron = 89, MF degree = 0.32, p-value = 0.003

**Fig. 2.** Example of single-faceted neurons. Each row shows the visualization of one neuron in grayscale and three different color computations.

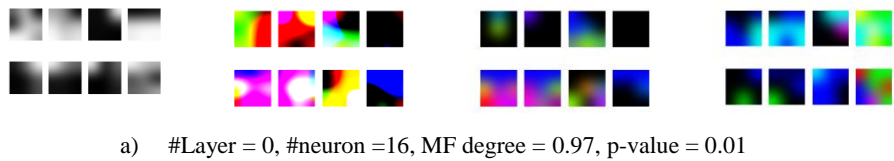

a)  #Layer = 0, #neuron =16, MF degree = 0.97, p-value = 0.01

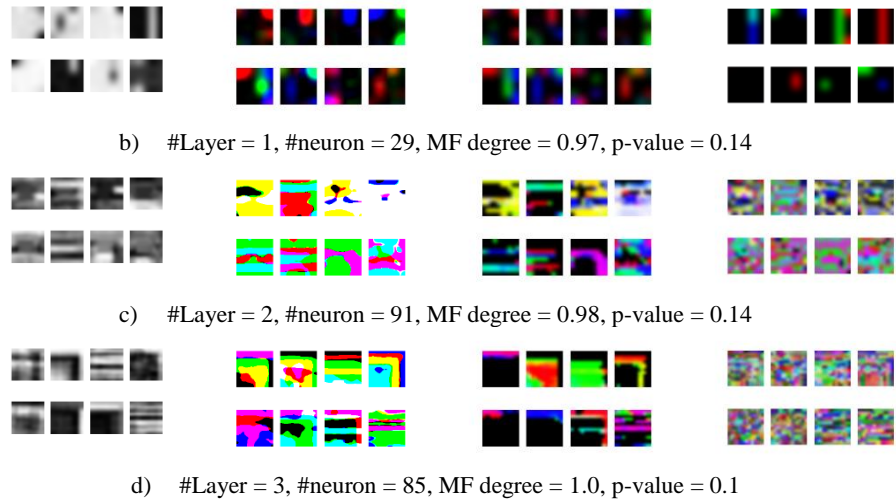

b)  #Layer = 1, #neuron = 29, MF degree = 0.97, p-value = 0.14

c)  #Layer = 2, #neuron = 91, MF degree = 0.98, p-value = 0.14

d)  #Layer = 3, #neuron = 85, MF degree = 1.0, p-value = 0.1

**Fig. 3.** Examples of multi-faceted neurons. Each row shows the visualization of one neuron in grayscale and three different color computations.

**SF Neurons**  **MF Neurons**

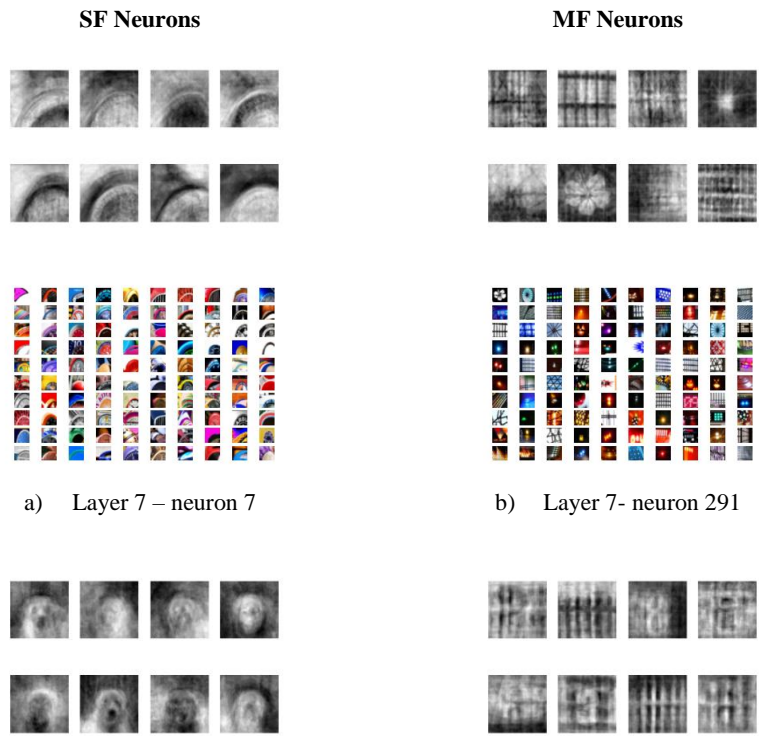

a)  Layer 7 – neuron 7     b)  Layer 7- neuron 291

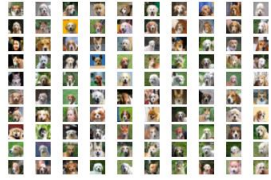
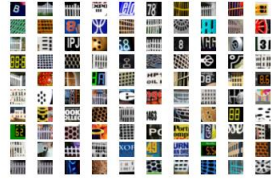

c) Layer 8 - neuron 0    d) Layer 8 - neuron 67

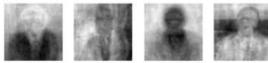
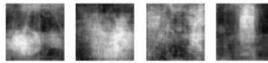

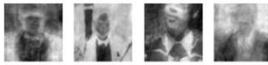
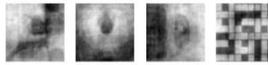

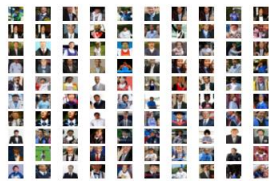
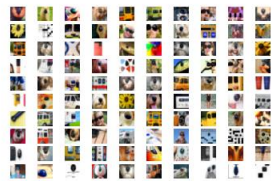

e) Layer 9 – neuron53    f) Layer 9 - neuron 297

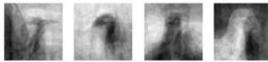
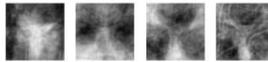

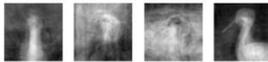
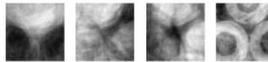

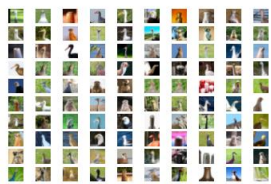
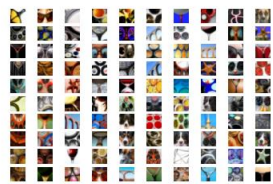

g) Layer 10 - neuron 185    h) Layer 10 - neuron 28

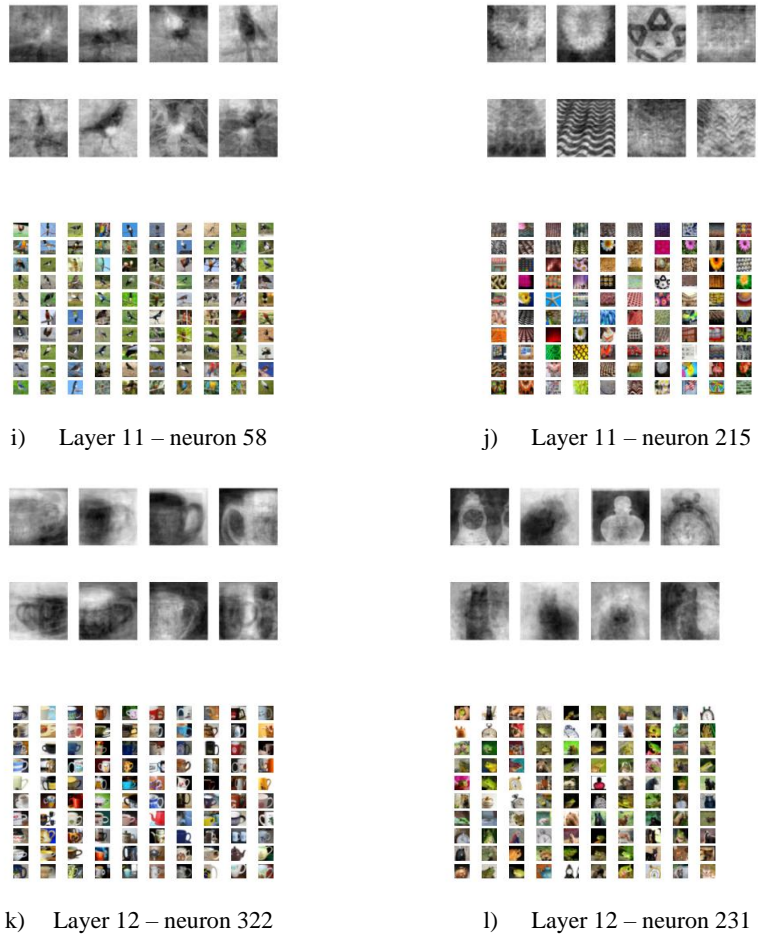

i) Layer 11 – neuron 58  j) Layer 11 – neuron 215

k) Layer 12 – neuron 322  l) Layer 12 – neuron 231

**Fig. 4.** Examples of SF and MF neurons and their corresponding top patches.

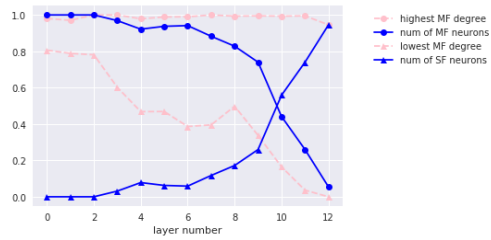

**Fig. 5.** Number of multi-faceted (MF) and single-faceted (SF) neurons and their corresponding MF degree at each deep layer

### 3.2 Neuronal similarity/dissimilarity matrices

In this section, the degree of pairwise neuronal similarity using the aforementioned conceptual encoding is computed. Generally, neuronal similarity can be measured from different perspectives. Neuroscientists tend to categorize neurons based on their morphological similarities. In this regard, they study neurons' shape, structure and connectivity (Stepanyants & Chklovskii, 2005)(Kong, Fish, Rockhill, & Masland, 2005). The association between morphology and functionality of neurons has also been investigated in (Stiefel & Sejnowski, 2007). In the present work, the similarity between trained neurons based on their response to images from different classes is investigated. To this aim, neurons similarity/dissimilarity matrices between all pairs of the CoF rows at each convolutional layer are constructed. The Pearson correlation is employed to measure the similarity, and Euclidean distance is applied to assess the dissimilarity between pairs of neurons in each layer. Fig. 6 presents the average conceptual similarity and dissimilarity of CoF matrices at each layer.

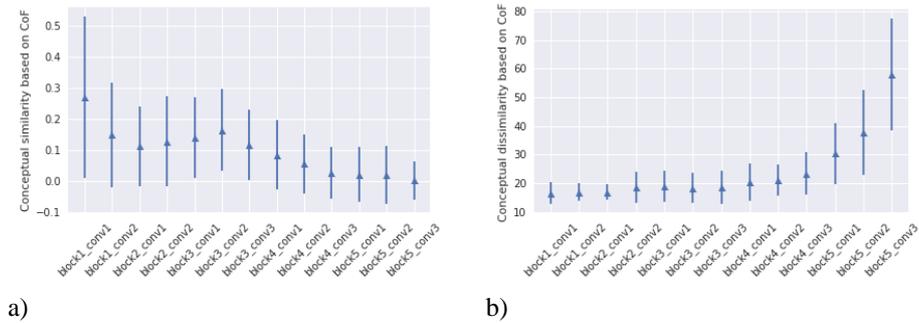

a)                                     b)

**Fig. 6.** Average values of a) Neuronal similarity, b) Neuronal dissimilarity

The results show that average conceptual similarities decrease as we go towards deeper layers. In other words, neurons become less correlated and more specialized in the higher layers. Whereas neuronal dissimilarities exhibit a more incremental progression. This can be explained by the fact that in the lower layers, neurons act as low feature detectors while neurons in the higher layers serve as shape-like detectors and so are more disposed to be less correlated. More importantly, the similarity matrices can also be viewed as a way of capturing the multi-faceted property. Higher similarities can be interpreted as a sign of multi-class preference. In contrary, low similarities highlight the existence of singular neurons that account for particular concepts. This interpretation is also compatible with finding in the previous section that the distribution of SF and MF neurons are not equal in different convolutional layers.

## 4 Conclusion

To recap, in this paper, neurons conceptual content in VGG16 pre-trained on ImageNet2012 data is studied. For this reason, a co-occurrence matrix is constructed to store the number of classes that activates each neuron. Furthermore, the multi-faceted degree is introduced and computed based on the measurement of the sparsity and flatness of each of the rows of the co-occurrence matrix. Besides, ICA technique is proposed as an approach for neuron visualization. The upside of taking this approach is that it can illustrate different facets of learned content of each neuron using a set of images. To my knowledge, this is the first study that puts forward the idea of generating multiple visualizations -in contrast to a single visualization for each neuron- to reflect the multi-tasking nature of neurons in a deep network. In fact, more investigations need to be carried out in future works in order to find alternative methods that can represent multiple functionalities of neurons. The experiments of this paper showed coherent and similar shapes in visualizations corresponding to SF neurons but diverse and divergent patterns in visualization of MF neurons. Finally, Neuronal similarities in each convolutional layer were computed based on their responses to each class of object. The aim of this experiment was to gain an understanding about neurons' pairwise functional similarity. The results demonstrate the unequal distribution of MF and SF neurons across deep layers. In essence, the results suggest that neurons that are lied in higher layers mirror less preference in responding to different classes, while neurons in the lower deep layers are more prone to be activated by dissimilar classes.

## Acknowledgment

This research has received funding from the European Union's Horizon 2020 research and innovation program under the Marie Skodowska-Curie grant agreement No 665919.